\documentclass[runningheads]{llncs}
\usepackage{graphicx}
\usepackage[utf8]{inputenc}
\usepackage{lscape}
\usepackage{makecell}

\begin{document}
\title{Spontaneous Theory of Mind for Artificial Intelligence}
\titlerunning{Spontaneous Theory of Mind for AI}
%
\author{Nikolos Gurney\inst{1}\orcidID{0000-0003-3479-2037} \and
David V. Pynadath\inst{1,2}\orcidID{0000-0003-2452-4733} \and
Volkan Ustun\inst{1}\orcidID{0000-0002-7090-4086}}
\authorrunning{N. Gurney et al.}
%
\institute{Institute for Creative Technologies, University of Southern California, 90094 USA 
\url{http://ict.usc.edu/}\\
\email{\{gurney,pynadath,ustun\}@ict.usc.edu}\and
Computer Science Department, University of Southern California, 90007 USA}

\maketitle
\begin{abstract}
Existing approaches to Theory of Mind (ToM) in Artificial Intelligence (AI) overemphasize prompted, or cue-based, ToM, which may limit our collective ability to develop Artificial Social Intelligence (ASI). Drawing from research in computer science, cognitive science, and related disciplines, we contrast prompted ToM with what we call spontaneous ToM---reasoning about others' mental states that is grounded in unintentional, possibly uncontrollable cognitive functions. We argue for a principled approach to studying and developing AI ToM and suggest that a robust, or general, ASI will respond to prompts \textit{and} spontaneously engage in social reasoning. 
\end{abstract}
\section{Introduction}

The ability to represent the content and state of each other's minds, commonly referred to as \textit{Theory of Mind} (ToM) \cite{premack1978does}, underpins much of human social cognition. It is theorized to take part in cognitive tasks as diverse as planning how to cheer somebody up who got stuck in traffic \cite{ho2022planning} and complex financial decision making \cite{de2013mind} (see also \cite{sutton1999bullying} in which they suggest its lack may be instrumental in contrasting behaviors, such as bullying). With much of human social cognition hinging on ToM, it is unsurprising that AI researchers see it as a possible solution to many hard problems in artificial social intelligence (ASI). Existing work in this space focuses on deliberate (prompted) rather than spontaneous (unprompted) ToM. In this theoretical contribution, we consider spontaneous ToM's role in humans and how a similar ability for AI could revolutionize work in ASI. 

The idea that people maintain internal representations of each other's minds is by no means new. Ancient philosophies, such as Platonism \cite{gerson2003knowing} and Confucianism \cite{pu2000confucius}, directly address social reasoning that matches modern definitions of ToM. It appears later during the Renaissance when philosophers, such as Descartes \cite{rozemond2009descartes}, considered the ability to represent other minds. And, of course, it has a place in contemporary research, much of which is founded on seminal work that examined primate ToM \cite{premack1978does}. The accepted definition for ToM, the ability to ascribe mental states to others and use those ascriptions for behavior predictions \cite{gurney2022robots}, stems from this research. The term \textit{theory of mind} comes from the idea that we develop an explicit theory as part of the underlying cognitive process for representing minds. This theory-theory is far from the only explanation. For example, other researchers have posited models that function based on simulations \cite{gordon1986folk} and that rely on the presence of a specific module for social cognition \cite{fodor1983modularity} (which may or may not rely on the development of explicit social theories). AI researchers have adapted many of these models for agents of all types, from virtual agents to robots \cite{gurney2021operationalizing,gurney2022robots}.

Extensive experimentation supports the various models and theories, which, at a high level, can be bifurcated into two categories: those that prompt ToM and those that do not. The quintessential ToM study, the Sally-Anne Test or False Belief Task (FTB), relies on a prompt: Participants observe a short skit that portrays a character, Sally, hiding a marble and another character, Anne, moving it in her absence. The key question, which is also a prompt to the participants, that tests ToM is, ``Where will Sally look for her marble?'' \cite{baron1985does,wimmer1983beliefs}. Many studies of ToM in human-computer interactions rely on the basic Sally-Anne test paradigm (e.g., \cite{pantic2003toward}). However, not all human cognition studies rely on prompts of ToM reasoning. For example, instead of asking participants for a particular action or explanation, researchers observe behaviors, such as gaze duration, to study \textit{spontaneous} ToM. A typical finding of these studies is that participants' gaze at a scenario is statistically longer when their beliefs differ from a character's (typically false) beliefs \cite{onishi200515,senju2009mindblind}. 

Actual spontaneous and prompted instances of ToM differ considerably from the common experimental paradigms. For example, while standing in line to purchase some goods, a person may spontaneously construct a model of the person in front of them which explains the items they are buying, e.g., they just reached a significant career milestone and are celebrating by treating themselves. Similarly, the person may have a friend whisper a prompt in their ear that triggers an entirely different model, e.g., ``That item is so indulgent, how wasteful!''  Despite this prompt, the observer may still come to the same conclusion. They also may conclude something more cohesive with their friend's suggestion. We believe there is more than a nuanced difference between these scenarios \textit{and} the difference matters to the development of artificial social intelligence capable of similar reasoning. The observer may not be another person but an AI-powered shopping assistant and the friend, the developer of the AI. In the former case, the shopping assistant might not say anything, but in the latter, it might chime in about budget constraints. The normative response hinges on the actual dynamics of the purchase event, but getting it wrong could sour a celebration or lead to an account overdraft. 

Our interpretation of the AI literature suggests that most ToM-enabled agents rely on models of prompted rather than spontaneous ToM. That is, they need explicit, pre-identified cues to know when they should activate their ToM. We remain agnostic about which approach is ``best'' or ``optimal,'' however, the overreliance on prompted ToM reasoning may leave AI researchers unable to definitively state whether a system does or does not have ToM---just like psychologists assumed that people with autism spectrum disorder did not have rich ToM until they started testing for spontaneous ToM \cite{senju2009mindblind}. Indeed, evidence of this is already emerging in the study of the social reasoning abilities of large language models \cite{kosinski2023theory,ullman2023large}. 

Based on our literature review, we lay out principles for studying ASI, take a stance on core features of robust (human-level) ASI, and highlight significant challenges to its realization. To better understand the current state-of-the-art in ASI, we need to start with a review of the psychological study of ToM. 

\section{Theory of Mind}
The ability to ascribe mental states to others and use those ascriptions for predicting or understanding behavior, i.e., Theory of Mind, is a core function of human social cognition. ToM helps us predict others' intentions and may play a critical role in developing and maintaining other cognitive abilities, such as emotional intelligence \cite{seidenfeld2014theory}. Moreover, its role is so prominent in cognition that we seemingly cannot help engaging it: People routinely ascribe mental states to inanimate objects, including computational systems, even when they are well aware of the objects' inanimacy \cite{epley2018mind}. This reality makes it unsurprising that we want intelligent systems to mimic, if not engage directly in, the same level of ToM reasoning as we do, i.e., have ASI. Considerable effort has gone into realizing this potential; an understanding of the modern research into ToM abilities will enable us to both frame the state-of-the-art ASI and explain why some researchers argue that building machines capable of thinking like humans necessitates engineering in intuitive theory usage, such as ToM \cite{lake2017building}.

\subsection{History}
Premack and Woodruff's research into chimpanzee social intelligence \cite{premack1978does} is commonly cited as the modern foundation of ToM research. They argue that the proper view of ToM is as a \textit{theory} because it involves making predictions about mental states that are not directly observable. Such prediction-making, they posit, requires a model founded on assumptions about how minds work. Diving into Premack and Woodruff's research will lead a reader back to famous research by Heider and Simmel that examines interpretations of apparent social interactions in non-human agents, specifically geometric shapes \cite{heider1944experimental}. Their experiment involved two-dimensional shapes that ``moved'' in and out of a rectangle, ``collided'' or ``interacted'' with each other, and, in so doing, displayed behavior that study participants interpreted as being social. Premack and Woodruff's research also has forward links to seminal work on the emergence of ToM in childhood development that gave us the prototype for empirical studies of ToM, the false belief test \cite{wimmer1983beliefs,baron1985does}. A cascade of research flowed from this early work, leading to a rich debate around the ToM cognitive process and how to study it. 

\subsection{Perspectives on the Theory of Mind Process}
Carving up the ToM body of research into meaningful divisions is not an easy task. Researchers, ourselves included \cite{gurney2021operationalizing}, typically divide the field based on distinct theoretical positions. A common starting point is to sort research into theory-theory and simulation-theory bins. Theory-theory states that people develop folk or naive theories to explain their social world through science-like experimentation that transpires during social interaction \cite{sellars1956empiricism,wimmer1983beliefs}. Simulation theory states that people accomplish ToM reasoning by simulating the cognitive state of others within their own mind \cite{gordon1986folk}. Each theoretical bin contains more precise theories describing the emergence and function of ToM. For example, the child-scientists perspective argues that the scientific method is a blueprint of ToM development and function \cite{gopnik1994minds,gopnik1999scientist}. 

Theories within both camps (and some that do not fit nicely into either) have contributed to advancing AI ToM \cite{gurney2021operationalizing,gurney2022robots}. However, we believe another distinction between approaches to studying ToM is equally meaningful to the development of ASI: ToM as a spontaneous mental process versus a prompted mental process. As will become clear, we use the term spontaneous to highlight the uncontrollable, often inexplicable, nature of thoughts related to the mental states of others. We shied away from the term unprompted because, we believe, it overemphasizes the role of otherwise unrelated stimuli in the ToM reasoning process. This logic is also why we landed on prompted: other terms, such as intentional or deliberate, fail to capture the role of external stimuli in the process. 

\subsubsection{Spontaneous Reasoning}
We define spontaneous reasoning as the set of unintentional mental processes that give rise to spontaneous thoughts. Spontaneous thoughts are those that happen seemingly uncontrollably and without reason \cite{morewedge2014perceived}. Spontaneous reasoning, albeit under various guises, has a storied history of scientific inquiry. After a long run of prompting study participants to explain the higher-order mental processes that gave rise to particular cognitive states, researchers began to question whether people actually have access to internal mental states at all \cite{nisbett1977telling}. Even though psychologists have gone to considerable lengths to refine their methods for eliciting cognitive processes from people (e.g., \cite{greenwald1995implicit,greenwald2017implicit}), it remains the case that there is no objective way to verify the veracity of reports on cognitive processes or know if a person is, in fact, accessing them \cite{ericsson2017protocol}. This methodological gap means that, at least given the current state of cognitive science, there is not a well-founded method for uncovering the spontaneous reasoning that underpins a given thought---including those related to the belief states of others.

Viewing ToM as spontaneous, i.e., a mental process grounded in unintentional, possibly uncontrollable cognitive functions, forces a reconsideration of existing theories, models, and empirical approaches. Consider theory-theory explanations. Having direct access to one's thoughts is implicit in the idea that children function as scientists experimenting with models of cognition to explain the reasoning of others. This implication exists because a person needs to isolate projections of others' mental states from their own mental states. Although they may not view it in this fashion, ToM researchers have acknowledged this reality, at least in their empirical methods: Children with greater inhibitory control of their cognition perform better in classic ToM tasks \cite{carlson2001individual}. This insight is an interpretation of an experimental phenomenon in which children report reality rather than what a person with a false belief believes. Given sufficient inhibitory control, a child can repress the urge to report on reality and provide the modal answer that indicates they ``have'' ToM. This observation does not give much insight into the cognitive functions underpinning the child's social intelligence. It just confirms that spontaneous reasoning can impact their social reasoning abilities. Moreover, it does not verify the emergence of ToM abilities, as they may have existed before inhibitory control was sufficient but hidden by the overwhelming impulse to report on reality.

\subsubsection{Prompted Reasoning}
Much like spontaneous reasoning, the psychology research community has long recognized the role of prompts, usually studied in the form of explicit questions or cues, in cognition. This recognition extends to the relatively small body of work committed to ToM, but it did go underappreciated for some time. Early work documenting the emergence of ToM capabilities noted how providing a mother with cues for reporting their child's speech abilities may have resulted in different responses. However, it did not consider how prompts to the children may impact utterances \cite{bretherton1982talking}. More significantly, prompts played a major role in research into the ToM abilities of autistic people. Much of this significant body of literature documents a high degree of correlation between deficits in ToM and autism spectrum disorder \cite{baron1985does,frith1994autism,happe1994advanced,perner1989exploration}, however, this conclusion is increasingly doubted \cite{gernsbacher2019empirical}, in part due to its reliance on explicit prompts. 

Referencing research from other domains of cognition can facilitate a better appreciation of prompts' possible impact on ToM. The role that prompts (and cues) can play in cognition is arguably understood thanks to research on memory. The long-established result in memory studies is that cues facilitate recall. For example, people are more likely to recall to-be-recalled words that are associated with cue words at the time of memory than to-be-recalled words without cue words or that were associated with the cue words after the initial memory event \cite{tulving1968effectiveness}. Endel Tulving later made the provocative argument that people do not actually forget some memories. Instead, the necessary triggering cue is lost \cite{tulving1974cue}. This insight spawned troves of research studying how exactly prompts impact, even alter, memory. Most relevant to the study of ToM is the finding that prompts can create false memories \cite{loftus2003make,loftus2005planting}. Imagining an experience, such as a trip to Disney Land, can lead a person to believe that some part of the imagined scenario actually took place \cite{braun2002make}. Another important insight is that the content of a cue can impact its effectiveness in generating the target memory \cite{uzer2017effect}.

It is reasonable to think of prompts as intentional, overt acts like giving a person a word to associate with a to-be-recalled word. In reality, prompts can lack intention and be covert. Asking a misbehaving child to consider how their actions may make another person act or feel is an example of an intentional, overt prompt intended to elicit a particular type of ToM reasoning. Overhearing two adults discuss the harm caused by another person's bad behavior may elicit the same ToM reasoning, but, in this case, the prompt was unintentional. In the case of ToM research, prompts that researchers unintentionally embedded in the experimental methods they used to study it, which we review below, contributed to them not accurately identifying when ToM emerges in childhood development. Once they recognized this oversight, researchers developed implicit ToM tasks to study spontaneous engagement in ToM reasoning.

\subsection{Empirical Approaches to Studying Theory of Mind}
Reviews of empirical approaches to studying ToM typically divide them into two categories: those with explicit and those with implicit requests for ToM reasoning, the latter of which researchers developed in response to criticisms about the former's ability to definitively detect the earliest examples of ToM in childhood \cite{low2012implicit}. The classic example of an explicit approach is the most well-known false belief test of ToM reasoning: the Sally-Anne test \cite{baron1985does,wimmer1983beliefs}. This test involves a small skit enacted by two dolls, Sally and Anne, followed by a simple question. Anne (and the child) see Sally hiding a marble in her basket and leaving the room. While Sally is away, Anne transfers the marble into her own box. Later, Sally returns to the room. A researcher then asks the child, ``Where will Sally look for the marble?'' 

With the impact of prompts on cognition in mind, the fatal critique of the Sally-Anne test is obvious: Asking a child where Sally will look for the marble may prompt them to consider the wider context of the skit. Not only does the child need to realize that Sally \textit{might} have a false belief (it may be the case Sally has a true belief because she knows Anne and predicted the stealing of the marble), but they need to recognize the demand of the researcher asking the question. Does the researcher expect a response demonstrating first-order beliefs (the modal response) or something richer? 

In response to this and other criticisms, researchers developed implicit ToM tasks that do not require direct interaction with research. For example, eye-tracking and measures of gaze duration suggest that children as young as fifteen months are able to predict the belief states of others \cite{onishi200515}. This and related experimentation build upon the violation-of-expectation method \cite{gergely1995taking}. The prototypical violation-of-expectation design has a child watch a person repeatedly reaching for one of two toys (the target and decoy) that are always in the same locations, in theory teaching the child of a preference for the target, after which the locations of the toys are switched. The person then randomly reaches for the target in its new location or the decoy that is now in the target's former location, with the result being that children tend to gaze longer at instances of decoy reaches \cite{woodward1998infants}. 

To elevate this design to a ToM test, Onishi and Baillargeon \cite{onishi200515} had infants watch an actor hide a toy in one of two locations. Next, its location was changed. For some infants, the actor observed the change, thus maintaining a true belief. For other infants, the actor did not observe the change, thus developing a false belief. The critical result confirming ToM is longer gaze times when the actor held a false belief (they did). This design, however, has also come under intense scrutiny given its less-than-ideal replication performance \cite{gernsbacher2019empirical,kulke2018implicit}. 

A meta-criticism of ToM research, thanks to the success and reality of the various criticisms of ToM tests, is now common in the literature: Popular explicit and implicit tests of ToM abilities may only capture other, low-level social-cognitive processes \cite{heyes2014submentalizing,quesque2020theory}. To illustrate, consider the dot perspective task \cite{samson2010seeing,santiesteban2014avatars}. Participants see one of four scenes: two dots are in front of an avatar (alternatively, a rectangle or arrow) in a room, one on the forward wall and another on the side wall, or the dot on the side wall is behind their field of view (or the location of the rectangle or arrow). Critically, faster correct responses are recorded when the arrow or avatar's perspective matches the participants (both dots are in front) \cite{santiesteban2014avatars}. When there is a rectangle in place of the arrow or avatar, there is no difference in response times \cite{samson2010seeing}. These combined results suggest that the dot perspective and related tasks are not studying Theory of Mind. Instead, they may only study other, low-level cognitive processes \cite{heyes2014submentalizing}. Summarily, there is not a widely accepted method for studying ToM, whether in children or adults, that lacks such fatal confounds. 

\section{Artificial Intelligence and Theory of Mind}
Note that this review is not an exhaustive or systematic review of AI ToM research. It is a high-level treatment of important work related to developing artificial social intelligence capable of ToM. Knowing the current state of affairs in the psychological study of ToM allows us to frame the accomplishments of artificial intelligence research related to ToM. Unlike human psychology, where ToM is studied for its own sake, in AI spaces, it is typically studied in applied settings as a means of overcoming another computational challenge. Thus, it is most present in fields like human-computer interaction \cite{gurney2021operationalizing} and human-robot interaction \cite{gurney2022robots}, but not as a fundamental problem of computation (although cognitive scientists have approached it as a computational problem). In the computational spaces that do pursue ToM research, much of it is focused on the human components of interactions, asking questions such as do people perceive ToM capabilities in the computational agents (whether it is there or not) (e.g., \cite{rilling2004neural} which documented similar neurological responses to human and computer opponents in the ultimatum game). 

Rudimentary examples of systems capable of modeling mental states have existed for nearly five decades (\cite{meehan1977tale}), but robust computational approaches to ToM only emerged early this century. Virtual agents that were increasingly human-like opened an opportunity for simulation work centered around predicting how people with differing perspectives and beliefs might interact with them. This research opportunity led to the development of PsychSim \cite{marsella2004psychsim}, a continually maintained platform that enables researchers to implement psychologically valid models of human behavior in virtual agents. The decision-theoretical capabilities of PsychSim allow it to model rich features of human cognition, including ToM \cite{pynadath2005psychsim}. Researchers have used PsychSim to study topics in human-computer/robot interactions as varied as calibrating trust \cite{wang2016trust}, disaster responses \cite{pynadath2023disaster}, and generating characters for interactive narratives that are capable of ToM \cite{si2014encoding,si2005thespian}.

Advances in machine learning (e.g., transformers \cite{vaswani2017attention}) and modern models of cognition (e.g., hierarchical Bayesian models \cite{tenenbaum2011grow}) have led to chatter about the possibility of developing human-level AI. Psychologists argue that in order to achieve this, computer scientists need to develop technologies beyond what is currently available \cite{cuzzolin2020knowing,lake2017building}. In response, computer scientists have demonstrated just how close to higher-order human cognition machines can come. Researchers from Google's Deepmind and Brain, for example, developed a ToM neural network model capable of not only learning to model the behavior of other artificial agents but also of passing classic ToM tests (e.g., false belief tests) \cite{rabinowitz2018machine}. 

More recently, large language models have overtaken research in artificial intelligence and related fields. Our interests lie in whether they, or any other computational models, can represent the belief and belief-like states of others, i.e., can they achieve a theory of mind? If we only consider their ability to perform in standard false belief tests, then the answer may be yes. A large-scale, multimodal language model outperformed six-year-old humans, achieving 75\% on 40 false belief tests \cite{kosinski2023theory}. Unfortunately for this model, trivial alterations to the tests undermine its performance \cite{ullman2023large}, and it seems reasonable to assume that these alterations would not impact human performance. Additionally, even when they are able to arrive at the correct answer to a false belief test, large language models appear to lack an understanding of why it is the modal answer \cite{trott2023large}, meaning they lack a ``theory'' of mind beyond pure statistical associations. Researchers argue that this shortcoming may result from a lack of pragmatic representation, a feature for which state-of-the-art models still rely on explicit definitions \cite{sap2022neural}.

Equally important to how researchers instantiate ToM in computational systems is how they test it---not unlike the case for humans. Most studies of computational ToM rely on some adaptation of a false belief test, and they are prompted studies---that is, the machines are explicitly asked to report on a hidden mental state. But, contingent on how one views training and validation, some approaches hint at spontaneous ToM. 

Obviously, how we ask questions of and give cues to machine intelligence is markedly different from how we do the same with humans. In many ways, the prompts given to AI models of ToM are present in their design, training, and tasks. To illustrate a design impact on ToM, consider PsychSim \cite{marsella2004psychsim,pynadath2005psychsim}. PsychSim agents rely on behavioral policies for both their own actions and interpreting the actions of other agents. These policies are typically a bounded lookahead procedure that accounts for other agents' actions as well as environmental dynamics. By adjusting the parameterization of these models, researchers can capture different ToM abilities. In practice, adjusting a model's lookahead length is telling it how to execute its ToM reasoning about other agents. In a human setting, it is akin to telling people in a prisoner's dilemma how many rounds they will play, knowledge that will undoubtedly alter how they model each other's decision-making. 

An alternative to endowing agents with predetermined models is allowing them to learn optimal parameterization from data \cite{wu2023multiagent}, or more generally, what model best accounts for variance in behavioral data \cite{rabinowitz2018machine}. ToMnet \cite{rabinowitz2018machine}, for example, posits an observer agent (which is a deep neural network) and policy-based agents. The latter execute their policies in simple grid-world environments, while the observer learns from their behavior. The observer has its own reward function which hinges on its ability to learn to represent multiple agents with differing policies, rewards, and parameterizations functioning in the grid-world. Critically, the observer agent has access during training and testing to the entire grid-world. This access allows it to embed end-to-end representations of behavior that it can later draw on when making ToM predictions. When considered in the context of the observer's reward function, this access is also explicitly prompting it about what to pay attention to. In effect, the observer only knows one type of question, so when presented with data, it answers that question (in this instance, what will a policy-based agent do next).

It is reasonable that computer scientists turn to psychology for ToM tasks. Human behavior, after all, is the source of our interest in ToM. Realizing human-level reasoning abilities in AI is the goal for many researchers, so naturally, they want to train, validate, and test their models just like psychologists test humans. Adopting test protocols from psychology, however, is a fraught practice. As noted, many of the tests are flawed. For example, the prototypical false belief test explicitly tells participants what behavior matters to the researcher and what answers are acceptable. In the case of humans, these details are implicitly negotiated through conversational norms, such as Gricean maxims \cite{grice1975logic}, that dictate how we interact. In the case of AI, it is often the case that the systems must be engineered to perform a given task. In so doing, the prompts, along with any flawed reasoning, are inbuilt. This reality is excellently illustrated between the two large language model papers we cited above: Kosinski gave the model standard tests of ToM and concluded that the system either had ToM or the test was flawed \cite{kosinski2023theory}. Ullman perturbed the tests in minor ways, which he argues would not prove a challenge for children, and found that the model failed and concluded that it did not have ToM \cite{ullman2023large}. 

AI systems that are able to perform on classic ToM tests still tend to be relatively narrow intelligences. Most state-of-the-art systems still rely on human-defined models of social reasoning. This reliance limits their abilities: current models of ToM lack generalizability, particularly when considered from a computational resources perspective \cite{gurney2021operationalizing}. AI ToM models are also constrained by the data that they learned from. As Ullman pointed out, this challenge is surmountable to some extent by adding new training cases \cite{ullman2023large}. Ultimately, however, that does not appear to create intelligence with the same abilities for generalization as humans, just one that has `seen' a lot of data, i.e., instances of a particular type of reasoning. Other domains, such as competitive video game models, more clearly illustrate this observation. Consider the state-of-the-art deep learning project AlphaStar, a model trained to excel in playing a particular game (StarCraft II) \cite{vinyals2019grandmaster}. Even though it is super-human in its abilities, humans who are able to generalize their strategies ultimately beat it. This illustration is analogous to the state of ToM and AI: although given enough data, models can perform at super-human levels, they remain brittle and are particularly prone to fail when confronted with tests that require generalization. 

We believe a reasonable hypothesis is that the prompt and prompt-like architectures researchers use when developing AI ToM models may inhibit generalization. Moving away from prompt-based and towards spontaneous reasoning models might uncover new insights into AI and human ToM. Artificial social intelligence, just like human intelligence, should be capable of more than just prompted social reasoning. State-of-the-art research, however, exclusively studies instances of prompted ToM. Assuming that ASI is the goal of studying AI ToM, which appears to be the case for many research teams, then a more principled approach to its study is needed.

\section{A Principled Approach to Studying Artificial Social Intelligence}
Theory of Mind is just one component of a broader class of cognitive skills that help people navigate the social world. Reproducing our ability to (seemingly) read each others' minds in a machine is an alluring prospect. Unfortunately, not only does it appear that currently available computational resources are not sufficient, but the psychology is also incomplete. These hurdles, of course, do not mean AI ToM and ASI are not worth pursuing. However, a more principled approach could benefit our collective efforts, particularly since we cannot rely entirely on human psychology research to provide definitive answers about the phenomena we are working to replicate.

\subsection{Consider how a Question Shapes the Answer}
There is an art to asking questions, whether of other people or AI. Social psychologists have long appreciated that how we ask a question can shape the answer we get from a respondent \cite{schwarz1999self}. This appreciation stems from the observation that answering a question requires an understanding of the semantic meaning of the actual words \textit{and} an appreciation of the intent of the questioner \cite{schober1992asking}. We believe the same general insight about asking questions applies to scientific research in that how a research question is asked can ultimately determine what answers are possible. For example, well-executed false belief tests can assess whether a child is capable of representing the belief state of another person, but does not directly answer the question of whether they have robust ToM and are capable of representing a rich set of mental states. The same is true of AI ToM. Research methods that rely on prompting for a response will generate answers that reflect the prompt(s), whereas research methods that rely on observation of spontaneous ToM will generate answers that reflect the structure of the observation. Both are valuable but can provide different answers about ToM. We believe that a foundational principle to any research is the consideration of how a question may shape the answer. Ullman has already demonstrated this in the LLM ToM space with his minor modifications to classic reasoning tasks that undermined the models' performance \cite{ullman2023large}.

\subsection{Focus on Definable Social Intelligence Skills}
The research into human ToM is undoubtedly informative (else we would not have reviewed it here), but its prominent role in how we think about ASI may hinder success. Theory of Mind takes more than assessing whether somebody has a false belief---beliefs are just one of many mental states we predict about each other. Nevertheless, belief plays an outsized role in the study of ToM abilities. Similar to beliefs are the general class of mental phenomena, sometimes called propositional attitudes, that reflect how we view a given proposition. These belief-like states are arguably as important to our social cognition but spectacularly underrepresented in the ToM literature. To illustrate, consider the statement \textit{student A is hoping for a good grade}, in which \textit{hoping} is the propositional attitude. Like a belief, another person can represent this attitude and hold their own attitude about it, thus have ToM related to the mental state of hoping. A robust ASI will keep track of more than just beliefs; arguably, it will be able to maintain representations of the same mental phenomena as humans, including beliefs, hopes, desires, and more. 

Theory of Mind eludes a consensus definition, possibly due to the related psychology research having a relatively narrow scope (i.e., focusing on beliefs rather than a more general ability). The lack of an accepted definition means that AI researchers are left to select, even invent, an operational definition of ToM that they can test their systems against. Formally defining the social intelligence skill in question is one principle we believe will yield quicker progress toward ASI. Formal definitions will facilitate such progress by improving our communication about the AI abilities we are working on and what we have accomplished. Formally defining the general ability to represent the mental states of others, i.e., ToM, and studying that definition in an AI fits with this principle. However, it may be more beneficial to define and study less nebulous abilities, for example, the ability to represent a belief, hope, or desire. This scope narrowing will naturally necessitate a change in how researchers talk about AI ToM, such as saying they are working on a specific ToM skill rather than ToM in general. The predicted benefit is that it may facilitate quicker, more effective development of ASI.

\subsection{Establish Ground Truth of Social Intelligence}
A valuable insight from existing tests of ToM, whether in psychology or computer science, is the need to rely on scenarios in which ground truth exists. Consider the Sally Anne test. Children likely view the prompt regarding the marble as genuine. That is, they do not think that the researchers are trying to deceive them or willfully withholding valuable information (e.g., Sally does not trust Anne based on prior marble-moving behavior). Meanwhile, adults might consider the prompt more cautiously. Although there is ground truth with respect to the marble, there is not necessarily ground truth available for Sally's belief state outside of what the researcher claims. This lack of ground truth is not problematic for naive agents. However, for sophisticated agents, it introduces another variable to consider. The agent, whether an adult or computational system, must consider other possibilities, such as the mental states of the person issuing the prompt and critical information about the characters they might not know. A wrong answer about where Sally will look for the marble may indicate robust ToM, just not in the expected fashion. 

\section{Robust Artificial Social Intelligence}
A robust ASI is an artificial intelligence with social reasoning abilities on par with adult humans. This definition differs from other robust AI systems in which the robustness typically describes their inertness to perturbations or adversarial attacks. Like humans, ASI may not always know the normative response to a social query or quandary. However, robust ASI will be able to reason about the mental states of the agents it observes and make predictions based on that reasoning. This ability means a robust ASI can respond to prompts about social situations, such as where a person will look for an object that they previously hid but was moved by another agent, as well as spontaneously engage in reasoning about the mental states of others, such as recognizing when a person is searching for something and offering to help. Minimally, robust ASI will:
\begin{itemize}
    \item \textbf{Respond to social prompts}: It will engage social intelligence to answer questions, react to cues, and respond to similar stimuli. 
    \item \textbf{Spontaneously engage in social reasoning}: It will engage in reasoning about the mental states of others without explicit or deliberate prompts. 
\end{itemize}
We think that the current models of Theory of Mind reasoning and associated empirical approaches to testing them do not fully consider both prompted and spontaneous reasoning. An over-emphasis on prompt-based tests of theories may falsely confirm their validity as models can learn the normative response to a prompt but not actually need to represent the mental state in question. Without an actual representation of the mental state, a model might fail to generalize its knowledge to trivially different social settings \cite{kosinski2023theory,ullman2023large}. 

\section{Conclusion}
Our collective, likely unintentional, focus on prompted Theory of Mind for AI will almost certainly slow and possibly inhibit progress toward ASI. The brilliant experimental paradigms that allowed researchers to unlock core insights about how people reason about the mental states of others, such as the Sally-Anne test, appear to have created this exact effect in human psychology. Relying on prompted studies of ToM is already leading to erroneous conclusions in AI research, as we saw with the competing studies of a large language model's ToM abilities. A more principled approach to studying AI ToM that considers how questions shape answers, focuses on definable social intelligence skills, and seeks out the ground truth of social intelligence, we believe, will enable the research community to avoid unnecessary delays in developing robust ASI that can respond to social prompts \textit{and} spontaneously engage in social reasoning. And, hopefully, such an approach will allow us to enjoy the help of an AI assistant that has enough social intelligence to not spoil the celebration of a small victory by reminding us of financial obligations. 

\subsubsection{Acknowledgements} The project or effort depicted was or is sponsored by the U.S. Government under contract number W911NF-14-D-0005. The content of the information does not necessarily reflect the position or the policy of the Government, and no official endorsement should be inferred

\subsubsection{Disclosures}
The authors have no competing interests to declare that are relevant to the content of this article.

\bibliographystyle{splncs04}
\bibliography{bib}

\begin{thebibliography}{10}
\providecommand{\url}[1]{\texttt{#1}}
\providecommand{\urlprefix}{URL }
\providecommand{\doi}[1]{https://doi.org/#1}

\bibitem{baron1985does}
Baron-Cohen, S., Leslie, A.M., Frith, U.: Does the autistic child have a “theory of mind”? Cognition  \textbf{21}(1),  37--46 (1985)

\bibitem{braun2002make}
Braun, K.A., Ellis, R., Loftus, E.F.: Make my memory: How advertising can change our memories of the past. Psychology \& Marketing  \textbf{19}(1),  1--23 (2002)

\bibitem{bretherton1982talking}
Bretherton, I., Beeghly, M.: Talking about internal states: The acquisition of an explicit theory of mind. Developmental psychology  \textbf{18}(6), ~906 (1982)

\bibitem{carlson2001individual}
Carlson, S.M., Moses, L.J.: Individual differences in inhibitory control and children's theory of mind. Child development  \textbf{72}(4),  1032--1053 (2001)

\bibitem{cuzzolin2020knowing}
Cuzzolin, F., Morelli, A., Cirstea, B., Sahakian, B.J.: Knowing me, knowing you: theory of mind in ai. Psychological medicine  \textbf{50}(7),  1057--1061 (2020)

\bibitem{de2013mind}
De~Martino, B., O’Doherty, J.P., Ray, D., Bossaerts, P., Camerer, C.: In the mind of the market: Theory of mind biases value computation during financial bubbles. Neuron  \textbf{79}(6),  1222--1231 (2013)

\bibitem{epley2018mind}
Epley, N.: A mind like mine: The exceptionally ordinary underpinnings of anthropomorphism. Journal of the Association for Consumer Research  \textbf{3}(4),  591--598 (2018)

\bibitem{ericsson2017protocol}
Ericsson, K.A.: Protocol analysis. A companion to cognitive science pp. 425--432 (2017)

\bibitem{fodor1983modularity}
Fodor, J.A.: The modularity of mind. MIT press (1983)

\bibitem{frith1994autism}
Frith, U., Happ{\'e}, F.: Autism: Beyond “theory of mind”. Cognition  \textbf{50}(1-3),  115--132 (1994)

\bibitem{gergely1995taking}
Gergely, G., N{\'a}dasdy, Z., Csibra, G., B{\'\i}r{\'o}, S.: Taking the intentional stance at 12 months of age. Cognition  \textbf{56}(2),  165--193 (1995)

\bibitem{gernsbacher2019empirical}
Gernsbacher, M.A., Yergeau, M.: Empirical failures of the claim that autistic people lack a theory of mind. Archives of scientific psychology  \textbf{7}(1), ~102 (2019)

\bibitem{gerson2003knowing}
Gerson, L.P.: Knowing Persons: a study in Plato. Clarendon Press (2003)

\bibitem{gopnik1994minds}
Gopnik, A., Meltzoff, A.N.: Minds, bodies, and persons: Young children's understanding of the self and others as reflected in imitation and theory of mind research.  (1994)

\bibitem{gopnik1999scientist}
Gopnik, A., Meltzoff, A.N., Kuhl, P.K.: The scientist in the crib: Minds, brains, and how children learn. William Morrow \& Co (1999)

\bibitem{gordon1986folk}
Gordon, R.M.: Folk psychology as simulation. Mind \& language  \textbf{1}(2),  158--171 (1986)

\bibitem{greenwald1995implicit}
Greenwald, A.G., Banaji, M.R.: Implicit social cognition: attitudes, self-esteem, and stereotypes. Psychological review  \textbf{102}(1), ~4 (1995)

\bibitem{greenwald2017implicit}
Greenwald, A.G., Banaji, M.R.: The implicit revolution: Reconceiving the relation between conscious and unconscious. American Psychologist  \textbf{72}(9), ~861 (2017)

\bibitem{grice1975logic}
Grice, H.P.: Logic and conversation. In: Speech acts, pp. 41--58. Brill (1975)

\bibitem{gurney2021operationalizing}
Gurney, N., Marsella, S., Ustun, V., Pynadath, D.V.: Operationalizing theories of theory of mind: A survey. In: AAAI Fall Symposium. pp. 3--20. Springer (2021)

\bibitem{gurney2022robots}
Gurney, N., Pynadath, D.V.: Robots with theory of mind for humans: A survey. In: 2022 31st IEEE International Conference on Robot and Human Interactive Communication (RO-MAN). pp. 993--1000. IEEE (2022)

\bibitem{happe1994advanced}
Happ{\'e}, F.G.: An advanced test of theory of mind: Understanding of story characters' thoughts and feelings by able autistic, mentally handicapped, and normal children and adults. Journal of autism and Developmental disorders  \textbf{24}(2),  129--154 (1994)

\bibitem{heider1944experimental}
Heider, F., Simmel, M.: An experimental study of apparent behavior. The American journal of psychology  \textbf{57}(2),  243--259 (1944)

\bibitem{heyes2014submentalizing}
Heyes, C.: Submentalizing: I am not really reading your mind. Perspectives on Psychological Science  \textbf{9}(2),  131--143 (2014)

\bibitem{ho2022planning}
Ho, M.K., Saxe, R., Cushman, F.: Planning with theory of mind. Trends in Cognitive Sciences  \textbf{26}(11),  959--971 (2022)

\bibitem{kosinski2023theory}
Kosinski, M.: Theory of mind may have spontaneously emerged in large language models. arXiv preprint arXiv:2302.02083  (2023)

\bibitem{kulke2018implicit}
Kulke, L., von Duhn, B., Schneider, D., Rakoczy, H.: Is implicit theory of mind a real and robust phenomenon? results from a systematic replication study. Psychological science  \textbf{29}(6),  888--900 (2018)

\bibitem{lake2017building}
Lake, B.M., Ullman, T.D., Tenenbaum, J.B., Gershman, S.J.: Building machines that learn and think like people. Behavioral and brain sciences  \textbf{40}, ~e253 (2017)

\bibitem{loftus2003make}
Loftus, E.F.: Make-believe memories. American Psychologist  \textbf{58}(11), ~867 (2003)

\bibitem{loftus2005planting}
Loftus, E.F.: Planting misinformation in the human mind: A 30-year investigation of the malleability of memory. Learning \& memory  \textbf{12}(4),  361--366 (2005)

\bibitem{low2012implicit}
Low, J., Perner, J.: Implicit and explicit theory of mind: state of the art.  (2012)

\bibitem{marsella2004psychsim}
Marsella, S.C., Pynadath, D.V., Read, S.J.: Psychsim: Agent-based modeling of social interactions and influence. In: Proceedings of the international conference on cognitive modeling. vol.~36, pp. 243--248 (2004)

\bibitem{meehan1977tale}
Meehan, J.R.: Tale-spin, an interactive program that writes stories. In: Ijcai. vol.~77, pp. 91--98 (1977)

\bibitem{morewedge2014perceived}
Morewedge, C.K., Giblin, C.E., Norton, M.I.: The (perceived) meaning of spontaneous thoughts. Journal of Experimental Psychology: General  \textbf{143}(4), ~1742 (2014)

\bibitem{nisbett1977telling}
Nisbett, R.E., Wilson, T.D.: Telling more than we can know: Verbal reports on mental processes. Psychological review  \textbf{84}(3), ~231 (1977)

\bibitem{onishi200515}
Onishi, K.H., Baillargeon, R.: Do 15-month-old infants understand false beliefs? science  \textbf{308}(5719),  255--258 (2005)

\bibitem{pantic2003toward}
Pantic, M., Rothkrantz, L.J.: Toward an affect-sensitive multimodal human-computer interaction. Proceedings of the IEEE  \textbf{91}(9),  1370--1390 (2003)

\bibitem{perner1989exploration}
Perner, J., Frith, U., Leslie, A.M., Leekam, S.R.: Exploration of the autistic child's theory of mind: Knowledge, belief, and communication. Child development pp. 689--700 (1989)

\bibitem{premack1978does}
Premack, D., Woodruff, G.: Does the chimpanzee have a theory of mind? Behavioral and brain sciences  \textbf{1}(4),  515--526 (1978)

\bibitem{pu2000confucius}
Pu, P.: From confucius to mencius: The confucian theory of mind and nature in the guodian chu slips. Contemporary Chinese Thought  \textbf{32}(2),  39--54 (2000)

\bibitem{pynadath2023disaster}
Pynadath, D.V., Dilkina, B., Jeong, D.C., John, R.S., Marsella, S.C., Merchant, C., Miller, L.C., Read, S.J.: Disaster world: decision-theoretic agents for simulating population responses to hurricanes. Computational and Mathematical Organization Theory  \textbf{29}(1),  84--117 (2023)

\bibitem{pynadath2005psychsim}
Pynadath, D.V., Marsella, S.C.: Psychsim: Modeling theory of mind with decision-theoretic agents. In: IJCAI. vol.~5, pp. 1181--1186 (2005)

\bibitem{quesque2020theory}
Quesque, F., Rossetti, Y.: What do theory-of-mind tasks actually measure? theory and practice. Perspectives on Psychological Science  \textbf{15}(2),  384--396 (2020)

\bibitem{rabinowitz2018machine}
Rabinowitz, N., Perbet, F., Song, F., Zhang, C., Eslami, S.A., Botvinick, M.: Machine theory of mind. In: International conference on machine learning. pp. 4218--4227. PMLR (2018)

\bibitem{rilling2004neural}
Rilling, J.K., Sanfey, A.G., Aronson, J.A., Nystrom, L.E., Cohen, J.D.: The neural correlates of theory of mind within interpersonal interactions. Neuroimage  \textbf{22}(4),  1694--1703 (2004)

\bibitem{rozemond2009descartes}
Rozemond, M.: Descartes's dualism. Harvard University Press (2009)

\bibitem{samson2010seeing}
Samson, D., Apperly, I.A., Braithwaite, J.J., Andrews, B.J., Bodley~Scott, S.E.: Seeing it their way: evidence for rapid and involuntary computation of what other people see. Journal of Experimental Psychology: Human Perception and Performance  \textbf{36}(5), ~1255 (2010)

\bibitem{santiesteban2014avatars}
Santiesteban, I., Catmur, C., Hopkins, S.C., Bird, G., Heyes, C.: Avatars and arrows: Implicit mentalizing or domain-general processing? Journal of Experimental Psychology: Human Perception and Performance  \textbf{40}(3), ~929 (2014)

\bibitem{sap2022neural}
Sap, M., LeBras, R., Fried, D., Choi, Y.: Neural theory-of-mind? on the limits of social intelligence in large lms. arXiv preprint arXiv:2210.13312  (2022)

\bibitem{schober1992asking}
Schober, M.F.: Asking questions and influencing answers. Questions about Questions: Inquiries into the Cognitive Bases of Surveys; Russell Sage Foundation: New York, NY, USA pp. 15--48 (1992)

\bibitem{schwarz1999self}
Schwarz, N.: Self-reports: How the questions shape the answers. American psychologist  \textbf{54}(2), ~93 (1999)

\bibitem{seidenfeld2014theory}
Seidenfeld, A.M., Johnson, S.R., Cavadel, E.W., Izard, C.E.: Theory of mind predicts emotion knowledge development in head start children. Early Education and Development  \textbf{25}(7),  933--948 (2014)

\bibitem{sellars1956empiricism}
Sellars, W., et~al.: Empiricism and the philosophy of mind. Minnesota studies in the philosophy of science  \textbf{1}(19),  253--329 (1956)

\bibitem{senju2009mindblind}
Senju, A., Southgate, V., White, S., Frith, U.: Mindblind eyes: an absence of spontaneous theory of mind in asperger syndrome. Science  \textbf{325}(5942),  883--885 (2009)

\bibitem{si2014encoding}
Si, M., Marsella, S.C.: Encoding theory of mind in character design for pedagogical interactive narrative. Advances in Human-Computer Interaction  \textbf{2014},  10--10 (2014)

\bibitem{si2005thespian}
Si, M., Marsella, S.C., Pynadath, D.V.: Thespian: Using multi-agent fitting to craft interactive drama. In: Proceedings of the fourth international joint conference on Autonomous agents and multiagent systems. pp. 21--28 (2005)

\bibitem{sutton1999bullying}
Sutton, J., Smith, P.K., Swettenham, J.: Bullying and ‘theory of mind’: A critique of the ‘social skills deficit’view of anti-social behaviour. Social development  \textbf{8}(1),  117--127 (1999)

\bibitem{tenenbaum2011grow}
Tenenbaum, J.B., Kemp, C., Griffiths, T.L., Goodman, N.D.: How to grow a mind: Statistics, structure, and abstraction. science  \textbf{331}(6022),  1279--1285 (2011)

\bibitem{trott2023large}
Trott, S., Jones, C., Chang, T., Michaelov, J., Bergen, B.: Do large language models know what humans know? Cognitive Science  \textbf{47}(7),  e13309 (2023)

\bibitem{tulving1974cue}
Tulving, E.: Cue-dependent forgetting: When we forget something we once knew, it does not necessarily mean that the memory trace has been lost; it may only be inaccessible. American scientist  \textbf{62}(1),  74--82 (1974)

\bibitem{tulving1968effectiveness}
Tulving, E., Osler, S.: Effectiveness of retrieval cues in memory for words. Journal of experimental psychology  \textbf{77}(4), ~593 (1968)

\bibitem{ullman2023large}
Ullman, T.: Large language models fail on trivial alterations to theory-of-mind tasks. arXiv preprint arXiv:2302.08399  (2023)

\bibitem{uzer2017effect}
Uzer, T., Brown, N.R.: The effect of cue content on retrieval from autobiographical memory. Acta psychologica  \textbf{172},  84--91 (2017)

\bibitem{vaswani2017attention}
Vaswani, A., Shazeer, N., Parmar, N., Uszkoreit, J., Jones, L., Gomez, A.N., Kaiser, {\L}., Polosukhin, I.: Attention is all you need. Advances in neural information processing systems  \textbf{30} (2017)

\bibitem{vinyals2019grandmaster}
Vinyals, O., Babuschkin, I., Czarnecki, W.M., Mathieu, M., Dudzik, A., Chung, J., Choi, D.H., Powell, R., Ewalds, T., Georgiev, P., et~al.: Grandmaster level in starcraft ii using multi-agent reinforcement learning. Nature  \textbf{575}(7782),  350--354 (2019)

\bibitem{wang2016trust}
Wang, N., Pynadath, D.V., Hill, S.G.: Trust calibration within a human-robot team: Comparing automatically generated explanations. In: 2016 11th ACM/IEEE International Conference on Human-Robot Interaction (HRI). pp. 109--116. IEEE (2016)

\bibitem{wimmer1983beliefs}
Wimmer, H., Perner, J.: Beliefs about beliefs: Representation and constraining function of wrong beliefs in young children's understanding of deception. Cognition  \textbf{13}(1),  103--128 (1983)

\bibitem{woodward1998infants}
Woodward, A.L.: Infants selectively encode the goal object of an actor's reach. Cognition  \textbf{69}(1),  1--34 (1998)

\bibitem{wu2023multiagent}
Wu, H., Sequeira, P., Pynadath, D.V.: Multiagent inverse reinforcement learning via theory of mind reasoning. arXiv preprint arXiv:2302.10238  (2023)

\end{thebibliography}

\end{document}